\documentclass{article}

\usepackage[final,nonatbib]{neurips_2018}

\usepackage[utf8]{inputenc} 
\usepackage[T1]{fontenc}    
\usepackage{hyperref}       
\usepackage{url}            
\usepackage{booktabs}       
\usepackage{amsfonts}       
\usepackage{nicefrac}       
\usepackage{microtype}      
\usepackage{multirow}
\usepackage{amsmath}

\usepackage{tikz}

\pgfmathdeclarefunction{gauss}{2}{%
  \pgfmathparse{1/(#2*sqrt(2*pi))*exp(-((x-#1)^2)/(2*#2^2))}%
}

\title{The effects of negative adaptation in Model-Agnostic Meta-Learning}

\author{
  Tristan Deleu \\
  Mila -- Universit\'{e} de Montr\'{e}al\\
  \texttt{tristan.deleu@gmail.com}
  \And
  Yoshua Bengio \\
  Mila -- Universit\'{e} de Montr\'{e}al\\
  CIFAR Senior Fellow
}

\begin{document}

\maketitle

\begin{abstract}
  The capacity of meta-learning algorithms to quickly adapt to a variety of tasks, including ones they did not experience during meta-training, has been a key factor in the recent success of these methods on few-shot learning problems. This particular advantage of using meta-learning over standard supervised or reinforcement learning is only well founded under the assumption that the adaptation phase does improve the performance of our model on the task of interest. However, in the classical framework of meta-learning, this constraint is only mildly enforced, if not at all, and we only see an improvement on average over a distribution of tasks. In this paper, we show that the adaptation in an algorithm like MAML can significantly decrease the performance of an agent in a meta-reinforcement learning setting, even on a range of meta-training tasks.
\end{abstract}

\section{Introduction}
\label{sec:introduction}
Humans are capable of learning new skills and quickly adapting to new tasks they have never experienced before, from only a handful of interactions. Likewise, meta-learning benefits from the same capacity of fast learning in the low-data regime. These algorithms are able to rapidly adapt to a new task through an \emph{adaptation phase}. In a method like Model-Agnostic Meta-Learning (MAML, \cite{DBLP:journals/corr/FinnAL17}), this phase corresponds to the update of the model's parameters. The main advantage of using meta-learning over standard supervised or reinforcement learning relies on the premise that this adaptation phase actually increases the performance of our model. However, in the usual meta-learning objective, there is no guarantee that the adaptation phase shows some improvement at the scale of an individual task.

Providing these kind of guarantees relates to safety concerns in reinforcement learning, where an agent not only optimizes its expected return, but with the additional constraint that this return must be above a certain threshold with high probability. Similarly, we would like to ensure that the performance of our agent does increase after adaptation, with high enough probability, over the whole distribution of tasks. In this paper, we show empirically that the adaptation phase in MAML can significantly decrease the performance on some continuous control problems in a meta-reinforcement learning setting, even on a range of meta-training tasks. Then inspired by the safety in reinforcement learning literature, we propose an alternative formulation of the meta-learning objective to encourage improvement over all tasks, and leave it as an open discussion.

\section{Related work}
\label{sec:related-work}

While meta-learning has been a long-standing problem in machine learning \cite{Thrun1998,bengio1992optimization,schmidhuber1987evolutionary}, there has been some fast progress in this field recently by combining ideas from meta-learning with deep-learning techniques. In the supervised setting, one of its major applications is in few-shot learning \cite{NIPS2016_6385,pmlr-v48-santoro16}, a regime that is generally out of reach of standard deep-learning techniques. There is similarly a long history of meta-learning applied to reinforcement learning problem \cite{hochreiter2001learning}. In the age of deep learning, meta-reinforcement learning has been approached from different angles, like optimization \cite{DBLP:journals/corr/WangKTSLMBKB16,DBLP:journals/corr/DuanSCBSA16} or with memory-augmented architectures \cite{Ritter360537}. The notion of fast learning and adaptation in meta-reinforcement learning can also be motivated from a neuroscience and psychology point of view \cite{Wang295964}.

In this paper, we are interested in particular to MAML \cite{DBLP:journals/corr/FinnAL17}, a model-agnostic algorithm that has shown success on both few-shot learning and meta-reinforcement learning. This algorithm has benefited from a lot of extensions, including ways to incorporate uncertainty estimates \cite{kim2018bayesian,finn2018probabilistic}. But besides simulated agents, MAML has also been applied to real-world environments, such as robotic tasks \cite{DBLP:journals/corr/abs-1709-04905}. This has been a key motivation for this work on the effect of the adaptation phase in MAML on the performance, where a lower performance might lead to some physical damage. This notion of a negative effect of adapting to a new task also relates to \emph{negative transfer} in the transfer learning literature \cite{rosenstein2005transfer}, where the transfer can hinder the performance of the agent.

\section{Model-Agnostic Meta-Learning in Reinforcement Learning}
\label{sec:maml-rl}

\subsection{Background}
\label{sec:background}

\paragraph{Reinforcement Learning} Even though meta-learning has been equally successful on both (few-shot) supervised and reinforcement learning problems, we only consider the \emph{meta-reinforcement learning} (meta-RL) setting in this paper. In the context of meta-RL, a task $\mathcal{T} = \langle\mathcal{S}, \mathcal{A}, p(\mathbf{s}'\mid \mathbf{s}, \mathbf{a}), r(\mathbf{s}, \mathbf{a})\rangle$ is defined as a Markov Decision Process (MDP), where we use standard notations from the reinforcement learning literature (see, for example, \cite{sutton1998reinforcement} for an introduction to reinforcement learning). For some discount factor $\gamma \in [0, 1]$, the return $G_{t}(\pi)$ at time $t$ is a random variable corresponding to the discounted sum of rewards observed after $t$, following the policy $\pi$:

\begin{equation}
  G_{t}(\pi) = R_{t+1} + \gamma R_{t+2} + \gamma^{2} R_{t+3} + \ldots = \sum_{k=0}^{\infty}\gamma^{k}R_{t+k+1}
  \label{eq:return}
\end{equation}

where $R_{t+1}$ is the (random) reward received after taking action $A_{t} = \pi(S_{t})$ in state $S_{t}$. 

\paragraph{Meta-Learning} Throughout this paper, we are working in a \emph{low-data regime}. In meta-RL, this corresponds to having access to a limited amount of interactions with the task of interest $\mathcal{T}$, of the order of $20$ trajectories in all of our experiments. Given this small dataset of trajectories $\mathcal{D}_{\mathcal{T}}$, the goal of the meta-learning algorithm is to produce a policy \emph{adapted} to this task, that is a policy that has a high expected return on $\mathcal{T}$.

\subsection{Model-Agnostic Meta-Learning}
\label{sec:maml}

In this work, we are interested in a meta-learning method based on \emph{parameter adaptation} and inspired by fine-tuning called \emph{Model-Agnostic Meta-Learning} (MAML, \cite{DBLP:journals/corr/FinnAL17}). The idea of MAML is to find a set of initial parameters $\theta$ of our (parametrized) policy $\pi_{\theta}$, such that only a single step of gradient descent is necessary to get new parameters $\theta_{\mathcal{T}}'$, where the corresponding policy $\pi_{\theta_{\mathcal{T}}'}$ is adapted to the task $\mathcal{T}$. More precisely, given a dataset $\mathcal{D}_{\mathcal{T}}$ of trajectories sampled from task $\mathcal{T}$, following the policy $\pi_{\theta}$, and a corresponding loss function $\mathcal{L}$, MAML returns new parameters $\theta_{\mathcal{T}}'$ defined as

\begin{equation}
  \theta_{\mathcal{T}}' = g(\mathcal{D}_{\mathcal{T}};\theta) = \theta - \alpha \nabla_{\theta}\mathcal{L}(\theta;\mathcal{D}_{\mathcal{T}})
  \label{eq:maml-update}
\end{equation}

where $\alpha$ is the step size for the gradient descent update. In meta-RL, this gradient update can be computed using vanilla policy gradient \cite{sutton1998reinforcement}, and $\mathcal{L}$ is typically a surrogate loss function like REINFORCE \cite{Williams92simplestatistical}, that can be defined as:

\begin{equation}
  \mathcal{L}(\theta;\mathcal{D}_{\mathcal{T}}) = -\frac{1}{N}\sum_{i=1}^{N}\sum_{t=0}^{\infty}\gamma^{t}\tilde{G}_{t}^{(i)}\log \pi_{\theta}(\mathbf{a}_{t}^{(i)}\mid \mathbf{s}_{t}^{(i)})
  \label{eq:reinforce-loss}
\end{equation}

where $\mathcal{D}_{\mathcal{T}} = \{(\mathbf{s}_{0}^{(i)}, \mathbf{a}_{0}^{(i)}, \mathbf{s}_{1}^{(i)}, \mathbf{a}_{1}^{(i)}, \ldots)\}_{i=1}^{N}$ and $\tilde{G}_{t}^{(i)}$ is the sample return at time $t$ of the $i\textsuperscript{th}$ trajectory. The overall meta-objective that is being optimized is an estimate of the generalization performance of the new policy $\pi_{\theta_{\mathcal{T}}'}$, over a distribution of tasks $p(\mathcal{T})$. This estimate can be built as the loss computed on some new dataset $\mathcal{D}_{\mathcal{T}'}$, sampled from task $\mathcal{T}$ following policy $\pi_{\theta_{\mathcal{T}}'}$:

\begin{equation}
  \min_{\theta} \mathbb{E}_{\mathcal{T}\sim p(\mathcal{T})}[\mathcal{L}(\theta_{\mathcal{T}}';\mathcal{D}_{\mathcal{T}}')] = \min_{\theta} \mathbb{E}_{\mathcal{T}\sim p(\mathcal{T})}[\mathcal{L}(g(\mathcal{D}_{\mathcal{T}};\theta);\mathcal{D}_{\mathcal{T}}')]
  \label{eq:outer-loss}
\end{equation}

\section{Negative adaptation}
\label{sec:negative-adaptation}

The minimization of the meta-objective in Equation~\eqref{eq:outer-loss} only encourages the adapted policy $\pi_{\theta_{\mathcal{T}}'}$ to have a high expected return, without any consideration of the policy $\pi_{\theta}$ we started with. There is no incentive for MAML to produce adapted parameters that improve the performance on the task of interest $\mathcal{T}$ over the initial policy. In particular, we could have a situation where the adaptation phase produces a policy whose return is lower than the initial policy. This kind of behavior could be critical in real-world environments, such as robotics, where we would like to trust the updated policy, without the need to systematically compare it to the policy we first gathered experience with. We call it \emph{negative adaptation}.

\subsection{Experimental Setup}
\label{sec:experimental-setup}

We evaluate the effects of negative adaptation in MAML on two continuous control problems based on the MuJoCo environments \cite{Todorov2012MuJoCoAP} Half-Cheetah and Ant. The experimental setup we use is identical to the one introduced in \cite{DBLP:journals/corr/FinnAL17}, which we recall briefly here. In order to build a variety of tasks based on a single environment, we changed the way the reward functions were computed, while keeping the dynamics fixed. In a first experiment, the reward function encourages the agent to move at a specific velocity $v_{goal}$, creating a one-to-one mapping between $v_{goal}$ and the tasks $\mathcal{T}$. In a second experiment, the reward function encourages the agent to either move forward or backward, effectively creating two different tasks. In all experiments, the expected values are similar to the ones reported in \cite{DBLP:journals/corr/FinnAL17}.

\subsection{Results}
\label{sec:results}

\begin{figure}[ht]
  \centering
  \includegraphics[width=\linewidth]{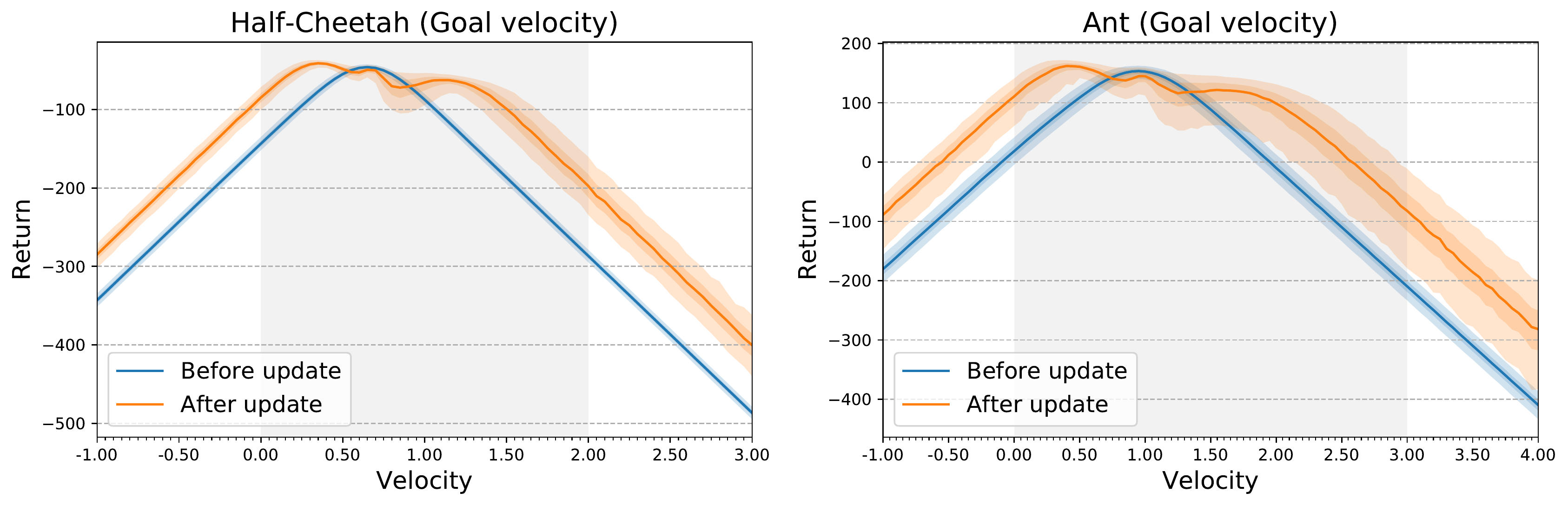}
  \caption{Sample returns before and after the one-step gradient update of MAML on two continuous control problems: (Left) Half-Cheetah and (Right) Ant, with goal velocity, for different values of the velocity (ie. different tasks). The median returns are shown in solid line, along with their $[25, 75]$ and $[5, 95]$ percentiles. The shaded region corresponds to the meta-training tasks range -- the tasks are sampled at meta-training with a uniform distribution over the corresponding velocities.}
  \label{fig:halfcheetah-ant-vel-returns}
\end{figure}

In Figure~\ref{fig:halfcheetah-ant-vel-returns}, we show the performance of the policies before and after the one step gradient update over a range of tasks, evaluated after meta-training. As expected, the updated policy generally gives a higher return than the initial policy, on both environments. Interestingly, this gap in performance seems to be mostly constant, even on tasks outside the meta-training range. This shows that MAML has some generalization capacity, even though the return decreases the further we are from the meta-training distribution. However, there is a range of tasks (velocities $[0.6, 1.1]$ for Half-Cheetah and $[0.6, 1.6]$ for Ant) on which we see the performance decrease significantly enough after adaptation.

\begin{figure}[ht]
  \centering
  \includegraphics[width=\linewidth]{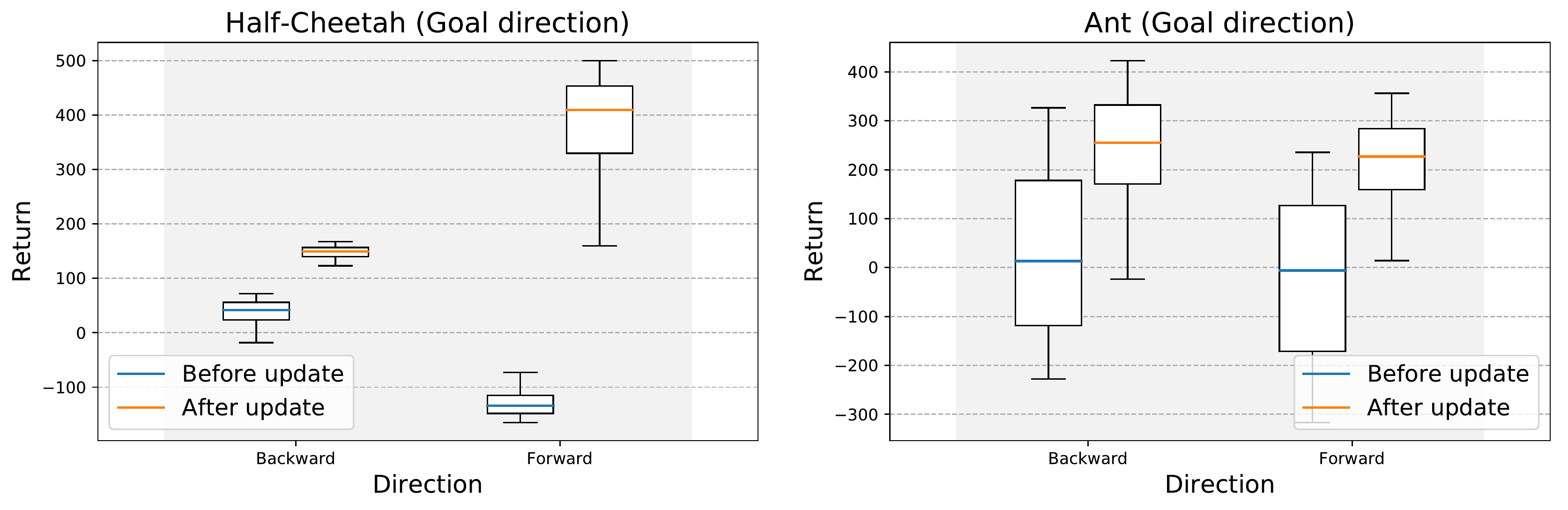}
  \caption{Sample returns before and after the one-step gradient update of MAML on two continuous control problems: (Left) Half-Cheetah and (Right) Ant, with goal direction. The boxplots show the median returns, along with their $[25, 75]$ and $[5, 95]$ percentiles. The shaded region corresponds to the meta-training tasks range.}
  \label{fig:halfcheetah-ant-dir-returns}
\end{figure}

The regime on which MAML shows negative adaptation seems to correspond to velocities where the return of the initial policy is already at its maximum. Intuitively, the meta-learning algorithm is unable to produce a better policy because it was already performing well on those tasks, leading to this decrease in performance. We believe that this overspecialization on some tasks could explain the negative adaptation. To illustrate this notion of overspecialization, we show the performance of MAML on continuous control problems with goal direction in Figure~\ref{fig:halfcheetah-ant-dir-returns}. While there is no severe negative adaptation on both of these problems, we can remark that the returns for the initial policy in the Half-Cheetah environment are heavily biased: the initial policy performs significantly better on the backward task (compared to the forward task), meaning that it specialized on this task. On the contrary, after the update, MAML shows little improvement on the backward task (the task it specialized on) compared to the forward task. This bias on the initial policy is nonexistent in the Ant environment, and shows equal improvement on both tasks after the adaptation.

\section{Discussion}
\label{sec:discussion}

In order to mitigate the effects of negative adaptation, we need to include a constraint on the improvement in our definition of the meta-objective in Equation~\eqref{eq:outer-loss}. While in Section~\ref{sec:negative-adaptation} the notion of improvement was not clearly defined, we wanted the return after the parameter update to be qualitatively higher than the return before the update. In order to characterize the improvement more precisely for a fixed task $\mathcal{T}$, we can introduce a random variable $\Gamma_{\mathcal{T}}(\theta)$, the difference between the returns of the policies before and after the parameter update:

\begin{equation}
  \Gamma_{\mathcal{T}}(\theta) = G_{0}(\pi_{\theta}) - G_{0}(\pi_{\theta_{\mathcal{T}}'})
  \label{eq:improvement}
\end{equation}

Avoiding negative adaptation for a specific task $\mathcal{T}$ translates to having $\Gamma_{\mathcal{T}}(\theta) \leq 0$, with high probability. Ideally, we would like to enforce this constraint over all tasks. However since we are working with a distribution over tasks $p(\mathcal{T})$, we can only guarantee this with high probability as well. Overall, we could define a new meta-objective as the following constrained optimization problem

\begin{equation}
  \min_{\theta} \mathbb{E}_{\mathcal{T}\sim p(\mathcal{T})}[\mathcal{L}(g(\mathcal{D}_{\mathcal{T}};\theta);\mathcal{D}_{\mathcal{T}}')] \qquad \textrm{s.t. } \Pr(\Pr(\Gamma_{\mathcal{T}}(\theta) \leq 0) \geq 1 - \beta) \geq 1 - \delta
  \label{eq:safe-optimization}
\end{equation}

with some values of $\beta, \delta \in (0, 1)$ that control the strength of the improvement constraint, at the level of a specific task for $\beta$, and globally over all tasks for $\delta$. Working with this kind of constrained objective relates to safety issues in reinforcement learning \cite{chow2014algorithms}. In safe RL, the objective is not only to maximize the expected return $\mathbb{E}[G_{0}(\pi)]$, but with guarantees on the individual values $G_{0}(\pi)$ may take. For example we could require the return to be lower bounded by some \emph{safe} value, with high probability. But despite their similarities, Equation~\eqref{eq:safe-optimization} actually differs from the standard framework of safe RL, making the optimization challenging. In early experiments on MAML with this modified meta-objective, we were not able to significantly reduce the effect of negative adaptation. Further work to build a more tractable approximation of Equation~\eqref{eq:safe-optimization} is currently ongoing.

\bibliography{references}
\bibliographystyle{plain}

\end{document}